\documentclass[a4paper, 10pt, conference]{ieeeconf}      
\usepackage{FG2027}
\usepackage{amssymb}
 \usepackage{amsmath}
\usepackage{algorithm}
\usepackage{algorithmic}
\usepackage{booktabs}
\usepackage{multirow}
\usepackage{graphicx}
\usepackage{url}
\usepackage{subcaption}

\FGfinalcopy 

\IEEEoverridecommandlockouts                              
\title{\LARGE \bf
Unlearning at Speed: Rapid Activation Dampening for Efficient Machine
Unlearning in Real-World Face Recognition
}

\author{\parbox{16cm}{\centering
    {\large Rouqaiah Al-Refai$^1$, Leo Scheubel$^2$, Philipp Terh\"{o}rst$^1$}\\
    {\normalsize
    $^1$ Johannes Gutenberg University Mainz, Germany\\
    $^2$ Paderborn University, Germany}} 
}

\begin{document}

\ifFGfinal
\thispagestyle{empty}
\pagestyle{empty}
\else
\author{Anonymous FG2027 submission\\ Paper ID \FGPaperID \\}
\pagestyle{plain}
\fi
\maketitle

\begin{abstract}

Machine unlearning has become increasingly important in face recognition, where regulations like the GDPR empower users to request the removal of their personal data from deployed models. However, most existing unlearning methods are designed for closed-set classification with small-scale models and training data, and therefore do not generalize to real-world face recognition systems, which operate in an open-set setting with large-scale models and data.
In this work, we propose Rapid Activation Dampening (RAD), a fast and targeted unlearning method tailored specifically for large-scale open-set face recognition. RAD approximates differences in average neuron activations between forget and retain identities, identifying and selectively suppressing neurons that are most associated with the targeted identity through a sensitivity-controlled dampening mechanism. 
We further propose the Forget Score (FS), which addresses the mismatch between existing unlearning metrics and open-set problems by providing a probabilistic interpretation of forgetting directly in the representation space.
Experiments across several recent large-scale face recognition architectures and datasets show that RAD achieves strong forgetting while preserving overall model utility and substantially improving unlearning speed. These findings establish RAD as a scalable and practical approach to machine unlearning in face recognition systems. Code will be made publicly available.
\end{abstract}

\section{Introduction}
As machine learning systems become embedded in critical infrastructure and everyday technologies, data privacy has become a central concern in research and regulation \cite{philipp2021Privacy}. Trained models often retain information about the individual data points, even when the original data has been deleted \cite{Carlini2019secret, ginart2019making}. This persistence challenges legal expectations for data removal under modern privacy regulations. Legal frameworks such as the General Data Protection Regulation (GDPR) enforce the right to erasure, requiring not only the deletion of data records but also the removal of their influence on model behavior "without undue delay" [9]. These obligations have motivated growing interest in machine unlearning, a research direction focused on removing the effect of specific training data without costly retraining models from scratch \cite{cao2015forget, ginart2019making}.

Face recognition (FR) represents a particularly urgent use case for unlearning. As one of the most widely deployed biometric technologies, it offers identity verification across various applications, from personal devices to surveillance and border control \cite{jain2011introduction}. These systems typically use deep neural networks \cite{taigman2014deepface, Schroff_2015_CVPR} trained on millions of face images \cite{MS_celeb_1m} to produce high-dimensional embeddings, which are stored and used for real-time recognition tasks. Once identity information is incorporated into such models, its removal becomes technically challenging.
Intuitively, this requires a regular retraining of the model from scratch on an updated training set that excludes the requested data, which is costly in terms of time, money and computational effort. Machine unlearning therefore aims to remove the influence of the data subject to a deletion request from the model more efficiently than full retraining.

However, most existing machine unlearning methods are tailored for closed-set classification tasks, trained and evaluated on comparably small-scale datasets compared to face recognition datasets, and applied to shallow models with relatively few layers and parameters. These settings do not hold in real-world face recognition systems. 
First, face recognition datasets comprise millions of images spanning hundreds of thousands of identities (e.g. MS1M \cite{MS_celeb_1m}), so the training set is orders of magnitude larger than those used in unlearning benchmarks (e.g. CIFAR-10 \cite{Krizhevsky2009LearningML} and CIFAR-100 \cite{Krizhevsky2009LearningML}). Second, the models are deep networks with tens of millions of parameters, whose behaviour cannot be characterized by the small, shallow architectures typically studied in machine unlearning. Third, training in face recognition relies on metric learning objectives \cite{Schroff_2015_CVPR, deng2019arcface, meng2021magface, DBLP:conf/eccv/BelabbaciWT26} rather than simple softmax classification, so the model is optimized to structure a high-dimensional embedding space instead of separating a fixed set of classes. Lastly, real-world face recognition is deployed in an open-set regime, in which the model must reject probes of identities never seen during training while accommodating newly enrolled identities without retraining, a setting that classifier-based machine unlearning does not address.

In this work, we propose Rapid Activation Dampening (RAD) to enable machine unlearning for open-set face recognition. RAD is a fast and scalable representation-level method for machine unlearning in face recognition systems.
The approach targets internal activations associated with requested identities for removal 
and adjusts them without requiring additional model retraining or access to classification outputs. It operates efficiently in large-scale embedding models and supports verification of forgetting quality while preserving model performance and significantly reducing unlearning time. Our contributions are summarized as follows:
\begin{itemize}
    \item \textbf{Rapid Activation Dampening:} We introduce a lightweight and scalable machine unlearning method for face recognition, called Rapid Activation Dampening (RAD), that selectively suppresses identity-relevant internal activations through a tunable dampening parameter derived from activation statistics between forget and retain identities. The method includes a hyperparameter to allow fine-grained control over the trade-off between unlearning speed and precision.
    \item \textbf{Forget Score:} Since standard closed-set unlearning metrics cannot be applied in open-set scenarios, we propose the Forget Score (FS). FS first measures the embedding shift between the model containing the forget identities and the model that has unlearned them, and then quantifies whether the shifts observed for the forget data and for the retain data originate from two distinct distributions. Together, this allow to quantify unlearning success in open-set scenarios.
\end{itemize}

\section{Related Work}
Machine unlearning has emerged as a critical area of research in response to growing privacy concerns and regulatory requirements such as the GDPR \cite{eu_gdpr_2016}. In deep learning, unlearning typically aims to remove the influence of undesirable data points from a trained model while preserving its utility on retained information without the cost of full retraining. The core objectives include computational efficiency, preserved performance on retained data, and verifiable forgetting of the targets \cite{Xu2023survey}. A straightforward approach is to remove the requested data from the training set and retrain the model from scratch \cite{Xu2023survey}. Although this provides a reference model trained without the forgotten data, it incurs substantial computational costs, particularly for large-scale deep learning systems.

To overcome this, retraining-based approaches were developed to localize forgetting to specific components or within specific stages of training. Bourtoule et al.~\cite{MU2021Bourtoule} introduced Sharded, Isolated, Sliced, and Aggregated (SISA) training, which partitions the training data into shards and slices to enable partial retraining for efficient unlearning. Graves et al.~\cite{graves2021amnesiac} proposed amnesic unlearning, which retrains the model after label corruption of the forget set, while Chundawat et al.~\cite{chundawat2023zero} introduced a student-teacher distillation strategy for selective knowledge erasure. Although effective, these methods are closely tied to classification-based architectures and become impractical in large-scale settings. However, retraining-free methods have gained attention due to their efficiency and model-agnostic nature. Notable examples include Selective Synaptic Dampening (SSD) \cite{Foster_Schoepf_Brintrup_2024} and its loss-free variant LFSSD \cite{LFSSD2024Foster}, which apply post hoc weight adjustments based on estimated parameters' importance. SSD typically computes parameter importance using the diagonal of the Fisher Information Matrix (FIM), which requires access to loss gradients and ground-truth labels; these resources are not available in pretrained face recognition models that rely on metric learning. 
LFSSD circumvents this limitation by replacing the Fisher-based estimation with a sensitivity-based measure that doesn't require loss or labels, making it adaptable to such settings. Accordingly, we consider SSD and LFSSD for our experimental comparison as recent, retraining-free approaches, with LFSSD building upon SSD through its loss-free parameter importance estimation. However, despite LFSSD's adaptability to embedding-based models, its computational cost remains a limitation when applied to large-scale face recognition models and datasets. This highlights the need for more computationally efficient alternatives, such as the approach proposed in this work.

Recent studies have investigated machine unlearning in face-related applications, including facial attribute classification and identity recognition. Choi et al.~\cite{choi2023machine} proposed benchmark datasets to study unlearning on face-related tasks such as age estimation and gender classification, which differ from identity recognition in both objective and architecture. Nair et al.~\cite{Nair2023selective} and Choi et al.~\cite{choi2024towards} introduced unlearning methods evaluated on facial data, however, their approaches rely on retraining within classification-based frameworks, often assuming softmax-based outputs.
These works do not address unlearning in real-world FR systems, where models operate in embedding space and are trained on large-scale identity datasets using metric learning. 
More recently, Zakharov et al.~\cite{Zakharov_2026_WACV} investigated identity unlearning in embedding-based face retrieval, proposing to disperse forgotten identity embeddings to prevent reliable re-identification while preserving retrieval performance for retained identities. Unlike approaches that optimize the embedding distribution through additional training, RAD directly modifies internal activation responses without requiring retraining, emphasizing computational efficiency and applicability to pretrained face recognition models.

In terms of evaluation, most existing unlearning methods rely on task-specific metrics such as accuracy drop on the forget set, retraining distance, or membership inference attack (MIA) success rate \cite{MIA2022Hu, MIA2017Shokri}. Other use classifier-based comparisons or proxy retraining classifiers to estimate residual influence \cite{MU2021Bourtoule}. While these metrics are suitable for classification tasks, they do not generalize naturally to models that output embeddings. In addition, techniques such as MIA are less effective for FR as they assume separated training, testing, and reference distributions, which is an assumption violated in practice due to overlapping in identity distribution across datasets.

\section{Proposed Unlearning Framework}

\begin{figure*}[ht]
    \centering
    \includegraphics[width=2.0\columnwidth]{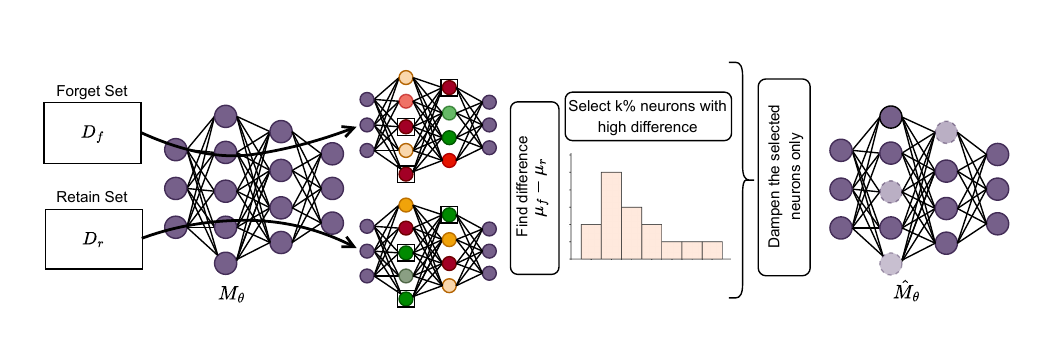}
    \caption{\textbf{Overview of the proposed unlearning framework}. The model $M$ processes samples from the dataset ($\mathcal{D}_f$ and $\mathcal{D}_r$) to compute neuron activation statistics at internal layers. Neurons with the highest positive activation differences for the forget set are selected and dampened to obtain the unlearned model $\hat{M}_\theta$. Neuron colors indicate relative activation strength, ranging from red (high activation) to green (low activation).}
    \label{fig:RAD}
\end{figure*}

\subsection{Problem Setting and Notation}
We address the problem of targeted identity unlearning in face recognition models, which aims to remove the influence of specific identities from a trained embedding model without retraining. Let $M_{\theta}: \mathcal{X} \rightarrow \mathbb{R}^{d}$ denote a face recognition model with frozen parameters $\theta$, mapping an input image $x \in \mathcal{X}$ to a feature embedding $z = M_{\theta}(x) \in \mathbb{R}^{d}$, where $d$ is the dimensionality of the output embedding space. The model is trained using any metric learning objective, such as triplet loss, ArcFace, or AdaFace loss \cite{Schroff_2015_CVPR, deng2019arcface, 9880230}, which enforces intra-class compactness and inter-class separation in the embedding space. In an unlearning scenario, given the original training dataset $\mathcal{D} = \{(x_{i}, y_{i})\}^{N}_{i=1}$, where $y_{i} \in \mathcal{Y}$ denotes the identity label, we consider a forget request that specifies a subset of class(es) $\mathcal{Y}_f \subset \mathcal{Y}$. The remaining classes form the retain set $\mathcal{Y}_r = \mathcal{Y} \setminus \mathcal{Y}_f$. Accordingly, the dataset is partitioned into the forget set as 
$\mathcal{D}_f = \{(x_i, y_i) \in \mathcal{D} \mid y_i \in \mathcal{Y}_f\}$ and the retain set as $\mathcal{D}_r = \{(x_i, y_i) \in \mathcal{D} \mid y_i \in \mathcal{Y}_r\}$. The goal is to modify $M_{\theta}$ to obtain a new model $\hat{M}_{\theta}$, such that for all $x \in \mathcal{D}_f$, the embeddings $\hat{z} = \hat{M}_{\theta}$ no longer preserve identity-specific information, while ensuring that embeddings for  $x \in \mathcal{D}_r$ remain close to the original, i.e., $\hat{M}_\theta(x) \approx M_\theta(x)$. Unlike classification networks, face recognition models do not rely on a final softmax layer to separate identities (e.g., triplet loss \cite{Schroff_2015_CVPR}). As a result, most existing machine unlearning methods, which operate by modifying or removing class-specific weights in the classifier, are not applicable in the FR setting. While it might be theoretically possible to adapt these methods by introducing proxy classifiers or auxiliary losses, doing so adds substantial tuning overhead and compromises the original embedding structure and performance, as shown in \cite{Nair2023selective}. Our approach avoids these complications by operating directly on the internal activations, enabling fast, tunable, and architecture-independent unlearning. Figure~\ref{fig:RAD} provides an overview of the proposed Rapid Activation Dampening (RAD) framework. Given a pretrained face recognition model and an identity targeted for unlearning, RAD estimates neuron importance by comparing activation statistics from the forget and retain sets. Neurons exhibiting the largest activation differences are selected and selectively dampened to reduce their contribution to the learned representations. The resulting model incorporates these modifications without requiring retraining or backpropagation.

\subsection{Activation-Based Neuron Selection}
To identify internal neurons most associated with the forget set, we analyze activations from internal layers of the model. Specifically, at each layer $l$, we focus on the activations immediately before the nonlinearity is applied. Let $\mathcal{A}_{l}(x) \in \mathbb{R}^{n_l}$ denote the pre-nonlinearity activation vector at layer $l$ for an input $x$, where $n_l$ is the number of activation units considered at that layer, with spatial dimensions aggregated where applicable. 
The process begins by computing the mean activation vector over the forget and retain sets:
\[\mu_{f}^{l} = \frac{1}{|{\mathcal{D}_{f}}|} \sum_{x \in \mathcal{D}_{f}} \mathcal{A}_{l}(x), \qquad  \mu_{r}^{l} = \frac{1}{|{\widetilde{\mathcal{D}}_{r}}|} \sum_{x \in \widetilde{\mathcal{D}}_{r}} \mathcal{A}_{l}(x) \] 
where $\widetilde{\mathcal{D}} \subset \mathcal{D}_r$ is a randomly sampled subset of retain-class examples, used to approximate the retain activation mean efficiently. We then compute the neuron-wise activation difference as $\Delta^{l} = \mu_{f}^{l} - \mu_{r}^{l}$, which reflects how much each neuron activates for forget-class samples compared to retain-class ones. Neurons with larger positive values for $\Delta^{l}$ are more responsive to the forget set and are thus prime candidates for suppression. To control the scope of the intervention, we rank the neurons according to their activation differences and select only the top $k\%$ of neurons with positive difference. These neurons form an index set $\mathcal{I}_{l} \subset \{1, \dots, n_l\}$, identifying the activation units selected for modification at layer $l$. The value of parameter $k$ controls the proportion of neurons modified, balancing the strength of forgetting against its effect on retained representations. This neuron selection strategy ensures that the most discriminative forget-relevant features are targeted for suppression, while minimizing unintended effects on the rest of the embedding space.

\subsection{Neuron Dampening via Scaled Offset}
Once the forget-relevant neurons have been selected, a dampening operation is applied to reduce their contribution to the final embedding. The goal is to suppress activations that respond disproportionately to forget-class inputs, without retraining the model. For each selected neuron $i \in \mathcal{I}$ at layer $l$, the dampening offset is computed as:
\[\delta_{i}^{l} = \gamma \cdot \Delta^{l}_{i} \]
where $\gamma \in [-1, 0]$ is a dampening strength parameter that determines how aggressively the neuron is suppressed. Larger negative values of $\gamma$ apply stronger dampening, while values closer to zero yield a weaker effect. The parameter thus offers a simple and intuitive way to control the dampening strength. The computed offset $\delta_i^l$ is applied directly to the activations of the selected neurons at layer $l$ during the forward pass as:
\[
\hat{\mathcal{A}}_l(x)_i = \mathcal{A}_l(x)_i + \delta_i^l.
\]

The dampening offsets are computed once per unlearning request and incorporated into the model, producing the unlearned model $\hat M_{\theta}$ without additional training or backpropagation.
Since $\Delta_i^l > 0$ for all $i \in \mathcal{I}$, the adjustment consistently reduces the activation of neurons that are most responsive to forget-class inputs, without altering the rest of the network. The overall approach introduces minimal computational overhead and provides fine-grained control over the unlearning behavior by adjusting $\gamma$ and the percentage $k$ of selected neurons. This approach allows class-specific identity information to be suppressed efficiently while preserving the generalization ability of the model on the retain set.

\subsection{Efficient Estimation for Retain-Class Activations}
In large-scale face recognition models, the retain set $\mathcal{D}_{r}$ may contain a vast number of samples, making it computationally expensive to evaluate all their activations when computing the retain-class activation mean, especially when handling forget requests consistently. To reduce this overhead, RAD approximates the retain activation mean using a randomly selected subset $(\widetilde{\mathcal D}_r\subseteq\mathcal D_r)$. 
The subset contains $10^{\mu}$ retain samples, where $\mu$ controls the trade-off between computational efficiency and unlearning performance.
This subset size can be adjusted according to available computational resources, allowing the reduction of importance estimation cost in restricted-resource scenarios.
Importantly, this approximation is restricted to the retain set, while the forget-set statistics are computed using all available forget samples, preserving the asymmetry of the unlearning formulation. 
The effect of $\mu$ on unlearning performance and computational efficiency is investigated experimentally. The full RAD unlearning procedure is summarized in Algorithm~\ref{alg:rad}.

\begin{algorithm}[tb]
\caption{Rapid Activation Dampening (RAD)}
\label{alg:rad}
\textbf{Input}: $M_\theta$, $\mathcal{D}_f$, $\mathcal{D}_r$, $l$ \\
\textbf{Parameter}: $k$, $\gamma$, $\mu$ \\
\textbf{Output}: $\hat{M}_\theta$
\begin{algorithmic}[1]
\STATE Randomly sample $\widetilde{\mathcal D}_r
\subseteq \mathcal D_r$ with
$|\widetilde{\mathcal D}_r|=10^\mu$
\STATE Compute $\mu_f^l \leftarrow \frac{1}{|\mathcal{D}_f|} \sum_{x \in \mathcal{D}_f} \mathcal{A}_l(x)$
\STATE Compute $\mu_r^l \leftarrow \frac{1}{|\widetilde{\mathcal{D}}_r|} \sum_{x \in \widetilde{\mathcal{D}}_r} \mathcal{A}_l(x)$
\STATE Compute $\Delta^l \leftarrow \mu_f^l - \mu_r^l$
\STATE Select index set $\mathcal{I}$ of top-$k\%$ neurons where $\Delta^l_i > 0$
\FOR{each $i \in \mathcal{I}$}
    \STATE Compute dampening offset $\delta_i^l \leftarrow \gamma \cdot \Delta^l_i$
    \STATE Apply offset $\delta_i^l$ to neuron $i$ at layer $l$
\ENDFOR
\STATE Save updated model as $\hat{M}_\theta$
\STATE \textbf{return} $\hat{M}_\theta$
\end{algorithmic}
\end{algorithm}

\section{Proposed Unlearning Metric}
\label{sec:forget_score}
\begin{figure}[ht]
    \centering
    \includegraphics[width=1.0\columnwidth]{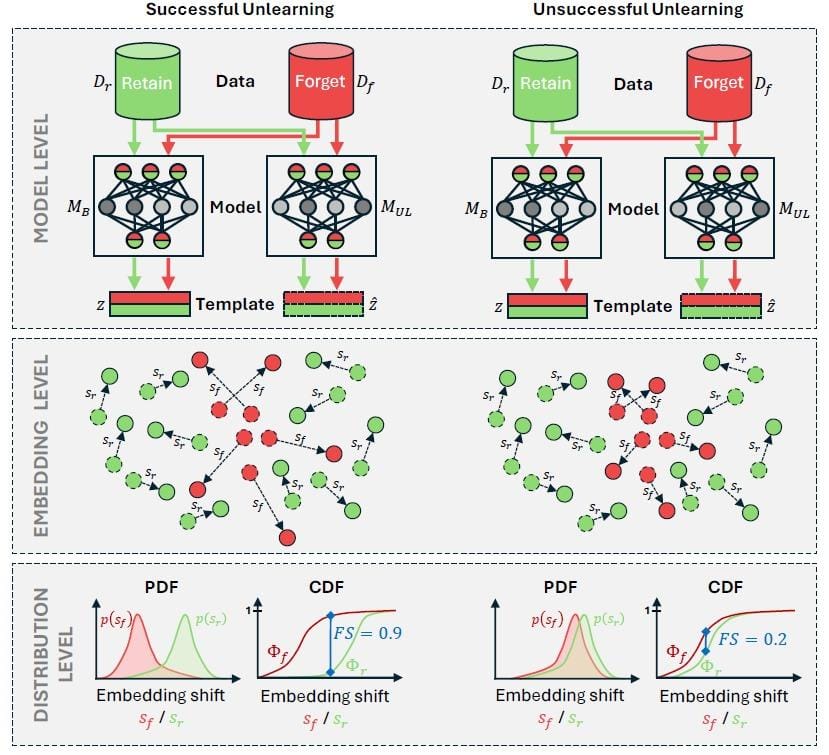}
    \caption{\textbf{Illustration of the Forget Score.} Successful unlearning shifts the forget-set similarity
distribution toward lower values while preserving
the retain-set distribution, resulting in a large
positive separation between their cumulative
distribution functions (CDFs) and a high FS.
Overlapping distributions yield a low FS.}
    \label{fig:forget_score}
\end{figure}

Identity information in open-set face recognition is encoded in feature embeddings, where representations of the same identity are expected to remain similar. Unlike classification-based models, where unlearning is commonly assessed through classification accuracy, face recognition models operate on embedding similarities without requiring a fixed set of classification labels. Consequently, evaluating identity unlearning requires examining changes in the representation space.
To capture these changes, we propose the Forget Score (FS), a metric that quantifies whether unlearning alters forgotten representations more strongly than retained ones. Rather than considering embedding changes independently, FS compares the changes in embeddings before and after unlearning across the forget and retain sets, assessing whether the forgotten identities experience greater representation shifts. The metric builds on the one-sided Kolmogorov-Smirnov (KS) statistic, which measures the maximum directional difference between two cumulative distributions. Figure \ref{fig:forget_score} illustrates the intuition behind the Forget Score by contrasting successful and unsuccessful unlearning, showing how differences in embedding shifts between forget and retain samples are reflected in their similarity distributions and corresponding cumulative distribution functions.

Let $z$ and $\hat{z}$ denote the embeddings produced for the same sample $x$ by the original and unlearned models, respectively. Since cosine similarity is commonly used to compare face representations, we employ it to quantify the directional change of each embedding after unlearning 

\[s(x) = \text{sim}(z, \hat{z}) = \frac{z \cdot \hat{z}}{\|z\|_2 \| \hat{z} \|_2} \]

A similarity close to one indicates that the embedding direction remains largely unchanged, whereas lower similarity indicates greater directional displacement. This allows us to assess how strongly unlearning affects individual representations without relying on classification outputs. The resulting similarity values form two sets,  $\mathcal{S}_f = \{s(x) \mid x \in \mathcal{D}_f \}$ and  $\mathcal S_r=\{s(x)\mid x\in \mathcal D^{FS}_r\}$, where $(\mathcal D^{FS}_r\subseteq\mathcal D_r)$ is a randomly sampled subset containing $10^\nu$ retain samples, with $\nu$ controlling the sampling size. This reduces the computational cost of FS evaluation on large-scale datasets. The sensitivity of FS to $\nu$ is discussed in the Results and Discussion section.

From these sets, we construct the empirical cumulative distribution functions (CDFs) $\Phi_{f}(s)$ and $\Phi_{r}(s)$, representing the proportions of forget and retain samples with cosine similarities at most (s). The Forget Score is then defined as:
\[\text{FS} = \sup_{s} \max(\Phi_{f}(s) - \Phi_{r}(s), 0)\]

The proposed formulation measures the maximum positive deviation between the two cumulative similarity distributions, specifically where the forget embeddings exhibit lower before-and-after similarity than retain embeddings, i.e., where $\Phi_{f}(s)>\Phi_{r}(s)$.  The directional nature of the metric ensures that only positive shifts in the desired direction are counted, while cases where the retain embeddings change more than those of the forget embeddings contribute nothing to the score. When FS equals zero, it indicates that no meaningful unlearning has occurred, or that retain-identity representations have been altered more than the forget identity. A higher FS reflects more effective forgetting, indicating that a larger fraction of forget-identity embeddings has been substantially altered while retain-identity embeddings remain relatively stable. Unlike existing classifier-dependent evaluation metrics relying on softmax outputs or retraining, FS offers a model-agnostic, efficient, and task-independent measure that operates directly on the representation space and reflects changes aligned with the unlearning objective. The full procedure is summarized in Algorithm~\ref{alg:fs}.

\begin{algorithm}[tb]
\caption{Forget Score (FS)}
\label{alg:fs}
\textbf{Input}: $M_\theta$, $\hat{M}_\theta$, $\mathcal{D}_f$, $\mathcal{D}_r^{\mathrm{FS}}$ \\
\textbf{Output}: $\mathrm{FS}$
\begin{algorithmic}[1]
\STATE Initialize empty sets $S_f \leftarrow \{\}$, $S_r \leftarrow \{\}$
\FOR{each $x \in \mathcal{D}_f$}
    \STATE Compute similarity $g \leftarrow \operatorname{sim}(M_\theta(x), \hat{M}_\theta(x))$
    \STATE Add $g$ to $S_f$
\ENDFOR
\FOR{each $x \in \mathcal{D}_r^{\mathrm{FS}}$}
    \STATE Compute similarity $g \leftarrow \operatorname{sim}(M_\theta(x), \hat{M}_\theta(x))$
    \STATE Add $g$ to $S_r$
\ENDFOR
\STATE Compute empirical CDFs $\Phi_f$ and $\Phi_r$ from $S_f$ and $S_r$
\STATE Compute $FS \leftarrow \sup_{\mathcal{S}} \max(\Phi_f(s) - \Phi_r(s), 0)$
\STATE \textbf{return} $\mathrm{FS}$
\end{algorithmic}
\end{algorithm}

\section{Experimental Setup}
\label{sec:setup} 
\subsection{Datasets and Models}
We conduct our experiments using two widely adopted face recognition models, ArcFace \cite{deng2019arcface} and AdaFace \cite{9880230}, both employing the iResNet50 backbone with approximately 43 million parameters. 
Unlike prior machine unlearning studies that evaluate primarily using small classification architectures, such as ResNet18, our experiments consider large-scale, embedding-based face recognition models.
These models produce identity-discriminative representations and enable face verification without relying on a classification head during inference. 
Evaluating both ArcFace and AdaFace further allows us to assess the applicability of unlearning methods across different face recognition training objectives.
We use pretrained recognition architectures trained on the MS1M-V2 \cite{MS_celeb_1m}, a large-scale face recognition dataset containing approximately 5.1 million images from over 85k identities.
In addition to serving as the original training source, it is used to construct the unlearning setting, where selected identities constitute the forget set $\mathcal{D}_f$, and the remaining identities form the retain set, $\mathcal{D}_r$.
This allows us to evaluate the identity-level unlearning directly on identities encountered during the original model training.
To assess the preservation of face recognition performance following unlearning, we evaluate the models on five standard face verification benchmarks, including LFW \cite{huang_inria-00321923}, AgeDB-30 \cite{moschoglou2017agedb}, CFP-FP \cite{cfpfp2016Sengupta}, CALFW \cite{abs-1708-08197} and CPLFW \cite{CPLFWTech}.
Following the standard verification protocols, we evaluate the predefined image pairs and report the mean verification accuracy. 
These benchmarks assess the robustness of face recognition under variations in age, pose, and unconstrained imaging conditions.

\subsection{Unlearning Protocol and Methods}
This paper focuses on identity-level unlearning in face recognition, where each unlearning request targets all training images associated with a specific identity. 
Following prior single-class unlearning evaluations \cite{LFSSD2024Foster, Foster_Schoepf_Brintrup_2024}, unlearning requests are formulated by selecting five distinct identities from the MS1M-V2 dataset and independently unlearned from the same initial pretrained model to avoid cumulative modifications. 
The same identities are used across all compared methods to ensure consistent evaluation, with results averaged over the five independent unlearning requests. The original pretrained model serves as the Baseline, providing a reference for recognition performance before unlearning. 
For the experiments, we adapted two recent unlearning approaches to the task of open-set recognition: SSD \cite{Foster_Schoepf_Brintrup_2024} and LFSSD \cite{LFSSD2024Foster}.

\textbf{SSD}, originally developed for classification-based unlearning, requires access to the model's classification outputs and corresponding loss for parameter importance estimation, which are not available in the standard open-set face recognition inference settings. To enable its application, the pretrained classification head is restored and combined with the face recognition backbone, allowing SSD to operate on the complete classification model. Parameter importance is estimated using the diagonal of the Fisher Information Matrix (FIM), computed separately over $\mathcal{D}$ and $\mathcal{D}_f$. Parameters exhibiting disproportionately high importance for $\mathcal{D}_f$ are then selectively dampened following the SSD \cite{Foster_Schoepf_Brintrup_2024} procedure.

\textbf{LFSSD}, also originally developed for classification models, it doesn't require a training loss for importance estimation, making it applicable to embedding-based face recognition. 
Accordingly, two configurations are considered. LFSSD-H incorporates the pretrained classification head with the backbone, providing a classification-based setting comparable to SSD. While LFSSD-E, excludes the classification head and applies the method directly to the embedding backbone, reflecting the standard inference configuration of face recognition systems. Both variants follow \cite{LFSSD2024Foster} importance estimation and dampening procedure, using $D$ and the targeted $D_f$.

\textbf{RAD}, in contrast, operates directly on the embedding backbone without requiring a classification head. Neuron importance is determined from activation statistics computed separately over $\mathcal{D}_f$ and $\mathcal{D}_r$. 
Unlike SSD and LFSSD, which rely on computationally heavy gradient-based parameter importance estimation, RAD uses fast forward-pass activation statistics to guide selective dampening in a rapid manner.

\subsection{Hyperparameter Tuning}
Hyperparameters are tuned independently for each unlearning method to account for differences in their importance estimation and dampening mechanisms.
For SSD and LFSSD-H, the hyperparameters $\alpha$ and $\lambda$ control parameters for selection and dampening, respectively. Specifically, $\alpha$ determines the threshold for identifying parameters important to the $\mathcal{D}_f$, while $\lambda$ regulates the magnitude of their dampening. Following the configurations reported for face unlearning in \cite{LFSSD2024Foster}, SSD is evaluated with $\alpha=50$ and $\lambda=0.1$, while LFSSD-H uses $\alpha=10$ and $\lambda=1.0$. Although these configurations were originally established for face classification tasks using a different dataset, they are adopted here for open-set face recognition on MS1M-V2. For LFSSD-E, an additional hyperparameter search is conducted over $\alpha \in \{1, 3, 5, 7, 10\}$ and $\lambda \in \{0.1, 0.5, 1.0\}$.
The original LFSSD configuration ($\alpha=10$, $\lambda=1.0$) is found to provide the best performance and is therefore retained. As for RAD the proportion of selected neurons, $k$, is varied over $\{0.25,0.5,1.0,2.0,5.0\}\%$, while the dampening strength, $\gamma$, is evaluated over $\{-0.1,-0.3,-0.5,-0.9\}$. Based on the trade-off between forgetting effectiveness and preservation of face verification performance, $k=0.5\%$ and $\gamma=-0.9$ are selected for the experiments.

\subsection{Unlearning Evaluation Metrics}
Unlike conventional classification-based unlearning, where forgetting is commonly evaluated using closed-set classification accuracy or Membership Inference Attack (MIA) based on classifier output, open-set face recognition relies on identity-discriminative embeddings rather than a fixed set of classification labels.
Consequently, these evaluation approaches are not directly applicable without additional adaptations. Instead, forgetting effectiveness is evaluated using the Forget Score, introduced in \ref{sec:forget_score}. Ideally, FS would be computed using the entire retain set; however, this can be computationally expensive for large-scale datasets. Therefore, we investigate the sensitivity of FS to retain-set sampling by varying the evaluation subset size as $10^\nu$, where $\nu \in \{2,3,4,5,6\}$.

To assess computational efficiency, we measure the unlearning runtime in seconds, excluding feature extraction and other preprocessing steps. We additionally report normalized runtime relative to RAD-100 for each backbone, allowing comparison of computational costs across different unlearning methods and sampling configurations. For RAD, we evaluate retain-set sampling configurations of $10^\mu$ samples, where $\mu \in \{2,3,4,5,6\}$, corresponding to retain subsets ranging from $10^2$ to $10^6$ samples, alongside the full retain set.

\section{Results and Discussion}
\begin{table*}[t]
\centering
\caption{Comparison of face verification performance, unlearning
effectiveness, computational efficiency, and applicability across
ArcFace and AdaFace. Verification accuracy is reported on five
standard benchmarks, while forgetting effectiveness is assessed
using the Forget Score. Unlearning runtime is reported in
seconds and normalized relative to RAD-100 for each backbone.
The final two columns indicate applicability to open-set recognition and model-agnostic operation.}
\label{tab:unlearning_comparison}
\setlength{\tabcolsep}{3.5pt}
\begin{tabular}{llcccccccccc}
\toprule
\textbf{Model} & \textbf{Method}
& \textbf{LFW} 
& \textbf{AgeDB-30}
& \textbf{CFP-FP}
& \textbf{CALFW}
& \textbf{CPLFW}
& \textbf{FS} $\uparrow$
& \textbf{Time (s)} $\downarrow$
& \textbf{Norm. Time} $\downarrow$
& \textbf{Open-set}
& \textbf{Model-agnostic} \\
\midrule
\multirow{10}{*}{ArcFace50}
& Baseline & 99.78 & 97.70 & 97.14 & 95.83 & 91.80 & -- & -- & -- & -- & -- \\
& SSD      & 99.78 & 97.71 & 97.08 & 95.85 & 91.77 & 0.95 & 30046.87 & 5533.49 $\times$ & $\times$ & $\times$ \\
& LFSSD-E    & 99.78 & 97.67 & 97.03 & 95.88 & 91.76 & 0.97 & 12954.11 & 2385.66 $\times$ & \checkmark & \checkmark \\
& LFSSD-H    & 99.78 & 97.67 & 97.05 & 95.82 & 91.74 & 0.95 & 15035.21 & 2768.92 $\times$ & $\times$ & $\times$ \\
& RAD-full      & 99.78 & 97.68 & 97.09 & 95.80 & 91.70 & 0.83 & 6109.96 & 1125.22 $\times$ & \checkmark & \checkmark \\
& RAD-1M      & 99.78 & 97.69 & 97.10 & 95.81 & 91.69 & 0.83 & 1460.63 & 268.99 $\times$ & \checkmark & \checkmark \\
& RAD-100K      & 99.78 & 97.68 & 97.08 & 95.80 & 91.69 & 0.83 & 150.68 & 27.75 $\times$ & \checkmark & \checkmark \\
& RAD-10K      & 99.78 & 97.68 & 97.05 & 95.81 & 91.72 & 0.82 & 19.77 & 3.64 $\times$ & \checkmark & \checkmark \\
& RAD-1K      & 99.78 & 97.67 & 97.05 & 95.82 & 91.74 & 0.82 & 6.69 & 1.23 $\times$ & \checkmark & \checkmark \\
& RAD-100      & 99.78 & 97.68 & 97.08 & 95.83 & 91.74 & 0.82  & 5.43 & 1.00 $\times$ & \checkmark & \checkmark \\
\midrule
\multirow{10}{*}{AdaFace50}
& Baseline & 99.60 & 97.88 & 97.07 & 95.98 & 91.68 & -- & -- & -- & -- & -- \\
& SSD      & 99.67 & 97.96 & 96.91 & 95.96 & 91.55 & 0.90 & 31792.70 & 5801.59 $\times$ & $\times$ & $\times$ \\
& LFSSD-E    & 99.69 & 98.00 & 97.03 & 96.06 & 91.57 & 0.97 & 13427.86 & 2450.34 $\times$ & \checkmark & \checkmark \\
& LFSSD-H    & 99.65 & 97.94 & 97.00 & 96.04 & 91.59 & 0.95 & 17152.76 & 3130.07 $\times$ & $\times$ & $\times$ \\
& RAD-full      & 99.63 & 97.92 & 97.04 & 96.03 & 91.54 & 0.85 & 7339.02 & 1339.24 $\times$ & \checkmark & \checkmark \\
& RAD-1M     & 99.63 & 97.92 & 97.04 & 96.03 & 91.54 & 0.85& 1571.78 & 286.82 $\times$ & \checkmark & \checkmark \\
& RAD-100K      & 99.63 & 97.92 & 97.04 & 96.03 & 91.54 & 0.85 & 161.94 & 29.55 $\times$ & \checkmark & \checkmark \\
& RAD-10K      & 99.61 & 97.94 & 97.03 & 96.02 & 91.52 & 0.85 & 20.93 & 3.82 $\times$ & \checkmark & \checkmark \\
& RAD-1K      & 99.61 & 97.96 & 97.00 & 96.01 & 91.52 & 0.84 & 6.89 & 1.26 $\times$ & \checkmark & \checkmark \\
& RAD-100      & 99.63 & 97.96 & 97.05 & 96.05 & 91.51 & 0.84 & 5.48 & 1.00 $\times$ & \checkmark & \checkmark \\
\bottomrule
\end{tabular}
\end{table*}

\subsection{Face recognition Performance}
We first evaluate the impact of unlearning on face recognition performance. Table \ref{tab:unlearning_comparison} reports the verification accuracy of the baseline model and the update models after unlearning across five standard benchmarks for ArcFace50 and AdaFace50. All face verification results are averaged over five selected forget identities, following previous works in reporting results on face datasets. 

Across both backbones, RAD maintains verification accuraccy close to that of the baseline model, regardless of the number of retain samples used for activation estimation. For ArcFace50, the largest absolute accuracy reduction across the five benchmarks is 0.11 percentage points, observed on CPLFW. Similarly, AdaFace50 exhibits a maximum reduction of 0.17 percentage points on CPLFW. The compared methods also preserve verification accuracy, with only minor variations relative to the pretrained baselines.

Notably, even the RAD variants using only $100$ or $1000$ retain samples maintain performance comparable to RAD performed with the full set, suggesting that estimating retain activation statistics from a relatively small subset is sufficient to preserve general face verification performance in the evaluated settings. These findings indicate that RAD can selectively modify representations associated with the forgotten identities while largely maintaining the discriminative capabilities of the pretrained models.

\subsection{Unlearning Effectiveness and Computational Efficiency}
\begin{figure}[t]
    \centering
    \begin{subfigure}[b]{0.45\textwidth}
        \centering
        \includegraphics[width=\linewidth]{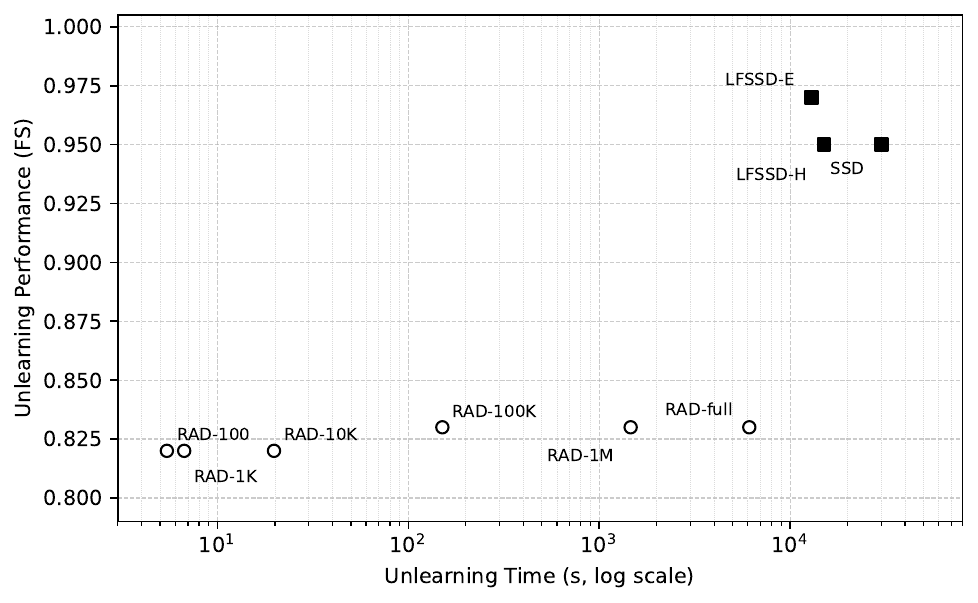}
        \caption{ArcFace50}
        \label{fig:arcface_fs_time}
    \end{subfigure}
    \hfill
    \begin{subfigure}[b]{0.45\textwidth}
        \centering
        \includegraphics[width=\linewidth]{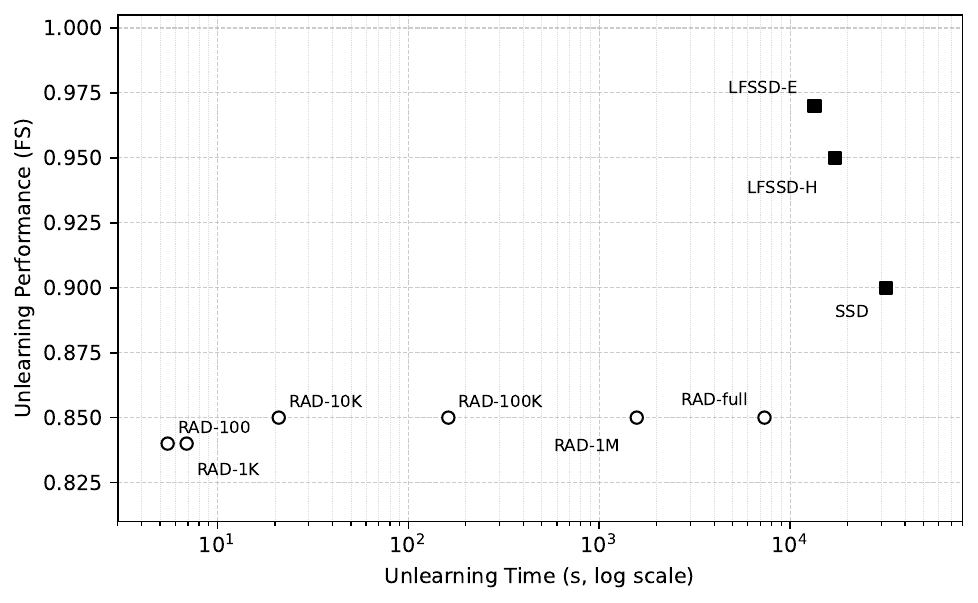}
        \caption{AdaFace50}
        \label{fig:adaface_fs_time}
    \end{subfigure}
    \caption{Unlearning performance versus runtime trade-off for
    ArcFace50 and AdaFace50. The figures compare SSD, LFSSD-E,
    LFSSD-H, and RAD specifically under different retain-set sampling configurations. Results are averaged across five forgotten identities. RAD achieves substantial reductions in unlearning time while maintaining a Forget Score close to its full-data configuration.}
    \label{fig:fs_time_tradeoff}
\end{figure}

We next evaluate unlearning effectiveness and computational efficiency using FS and unlearning runtime. 
As reported in Table \ref{tab:unlearning_comparison}, RAD achieves Forget Scores ranging from 0.82 to 0.83 for ArcFace50 and from 0.84 to 0.85 for AdaFace50. While LFSSD-E achieves a higher FS of 0.97 on both backbones, it requires approximately twice the unlearning time of RAD-full and over 2,300 times that of RAD-100. This highlights RAD's advantage in achieving substantial differential representation changes within considerably shorter execution times while preserving face recognition performance.

An important distinction among the compared methods concerns their applicability to different face recognition training objectives. Classification-dependent approaches, such as SSD and LFSSD-H in our adaptations, require access to classification-related information, which may not be available for models trained using metric learning objectives such as the triplet loss employed by FaceNet. In contrast, RAD operates on internal activation statistics without requiring a classification head or a specific training loss, making it applicable to embedding-based face recognition models independently of their original training objectives. LFSSD-E similarly avoids reliance on a classification head by operating directly on the learned embeddings. However, its high computational cost remains a limitation, particularly when dealing with large-scale face recognition models and datasets. This distinction is particularly relevant to open-set face recognition, where recognition is performed through embedding similarities rather than predictions over a fixed set of training identities.

Beyond its applicability, the main advantage of RAD lies in its computational efficiency. As shown in Table \ref{tab:unlearning_comparison}, RAD-full completes unlearning in approximately 6,110 seconds for ArcFace50 and 7,339 seconds for AdaFace50, requiring less time than SSD and both LFSSD adaptations. More importantly, estimating retain activation statistics from a subset substantially reduces the computational overhead. With only 10,000 retain samples, RAD completes unlearning in approximately 20 seconds for either backbone, representing a reduction of more than 99\% in runtime compared with RAD-full, while maintaining similar FS and verification accuracy.

Figure \ref{fig:fs_time_tradeoff} further illustrates the trade-off between unlearning effectiveness and computational efficiency for ArcFace50 and AdaFace50. Across both backbones, RAD achieves relatively stable FS values while substantially reducing unlearning time through an effective estimation of the retain-set. As reported in Table \ref{tab:unlearning_comparison}, RAD-100K requires only 27.75$\times$ and 29.55$\times$ the runtime of RAD-100 for ArcFace50 and AdaFace50, respectively, compared with 1,125.22$\times$ and 1,339.24$\times$ for RAD-full. More notably, RAD-10K achieves FS values of 0.82 and 0.85 with normalized runtimes of only 3.64$\times$ and 3.82$\times$, respectively. In contrast, LFSSD-E requires 2,385.66$\times$ and 2,450.34$\times$ the runtime of RAD-100. Thus, RAD-10K is approximately 655$\times$ and 642$\times$ faster than LFSSD-E on ArcFace50 and AdaFace50, respectively, while maintaining face verification performance close to the original models. These findings demonstrate that RAD can achieve consistent unlearning effects with substantially reduced computational requirements, without relying on exhaustive retain-set activation estimation.

Although LFSSD-E achieves higher FS values of 0.97 on both backbones, this improvement comes with considerably greater computational requirements. RAD therefore offers a different balance between unlearning effectiveness and efficiency, emphasizing rapid targeted interventions while preserving face recognition performance. Such a trade-off is particularly relevant to practical deployments, where deletion requests may be issued repeatedly rather than processed as isolated operations. Investigating the cumulative effects of sequential unlearning requests remains an important direction for future work.

\subsection{Forget Score and Sampling Analysis}
\begin{figure}[t]
    \centering
    \includegraphics[width=\linewidth]{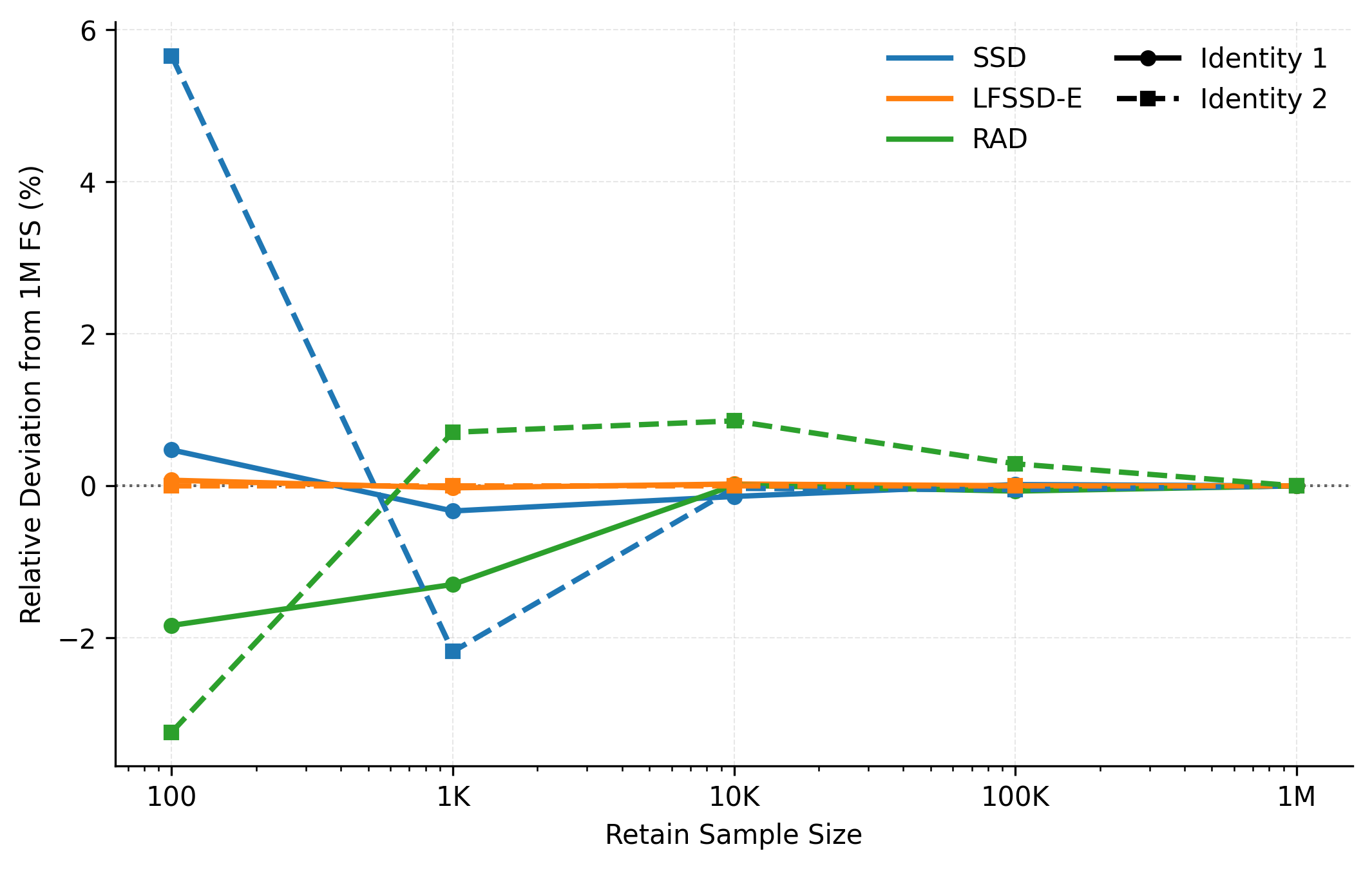}
    \caption{Stability of the Forget Score (FS) across different retain sample sizes. Relative deviations are computed with respect to the FS obtained using 1M retain samples, independently for each method and forgotten identity.}
    \label{fig:fs_sampling_stability}
\end{figure}

To investigate the sensitivity of the proposed Forget Score metric to the number of retain samples used during evaluation, we compute FS using retain subsets of increasing sizes, ranging from $10^2$ to $10^6$ samples. Unlike the sampling employed during RAD unlearning, which approximates retain activation statistics, this analysis examines the stability of FS estimation itself. Figure \ref{fig:fs_sampling_stability} presents the resulting scores for two forgotten identities across SSD, LFSSD-E, and RAD.

The results show that FS estimates can vary when relatively small retain subsets are used, with the extent of variation depending on both the unlearning method and the forgotten identity. For identity 1, the scores remain largely stable across the evaluated sample sizes, particularly for LFSSD-E and SSD, while RAD exhibits a small increase before stabilizing. In contrast, identity 2 exhibits greater sensitivity to the retain sample size, particularly for SSD and RAD. For SSD, FS decreases from approximately 0.93 at $10^2$ samples to 0.86 at $10^3$, before stabilizing near 0.88 for larger subsets. Similarly, RAD increases from approximately 0.84 at $10^2$ samples to 0.87 at $10^3$, followed by relatively small variations as the sample size increases. LFSSD-E remains consistently close to 1.0 across both identities and all evaluated sample sizes.

Overall, the results indicate that increasing the retain evaluation subset size improves the stability of FS estimates, with relatively limited variation observed at larger sample sizes. In particular, the scores obtained using $10^4$, $10^5$, and $10^6$ retain samples are generally comparable, suggesting that evaluating the complete retain dataset is not necessary to obtain a reasonably stable FS estimate in the examined cases. Accordingly, we adopt $10^4$ retain samples for FS computation in our experiments, balancing estimation stability and computational efficiency.

\section{Conclusion}
In this work, we introduced Rapid Activation Dampening (RAD), a retraining-free approach for identity unlearning in open-set face recognition. By selectively modifying internal activations and leveraging retain-set sampling, RAD enables efficient unlearning while preserving recognition performance. We further propose the Forget Score (FS), an embedding-based metric for evaluating the differential impact of unlearning on forgotten and retained representations. Experiments across different face recognition backbones demonstrate that RAD substantially reduces computational costs while maintaining competitive unlearning effectiveness and stable verification performance. Our analysis further highlights the benefits of retain-set sampling for efficient unlearning and FS evaluation, supporting the potential of RAD for scalable identity unlearning.


{\small
\bibliographystyle{ieee}
\bibliography{egbib}

@ARTICLE{philipp2021Privacy,
  author={Meden, Blaž and Rot, Peter and Terhörst, Philipp and Damer, Naser and Kuijper, Arjan and Scheirer, Walter J. and Ross, Arun and Peer, Peter and Štruc, Vitomir},
  journal={IEEE Transactions on Information Forensics and Security}, 
  title={Privacy–Enhancing Face Biometrics: A Comprehensive Survey}, 
  year={2021},
  volume={16},
  number={},
  pages={4147-4183},
  doi={10.1109/TIFS.2021.3096024}
}

@inproceedings{Carlini2019secret,
author = {Carlini, Nicholas and Liu, Chang and Erlingsson, \'{U}lfar and Kos, Jernej and Song, Dawn},
title = {The secret sharer: evaluating and testing unintended memorization in neural networks},
year = {2019},
isbn = {9781939133069},
publisher = {USENIX Association},
address = {USA},
booktitle = {Proceedings of the 28th USENIX Conference on Security Symposium},
pages = {267–284},
numpages = {18},
location = {Santa Clara, CA, USA},
series = {SEC'19},
}

@inbook{ginart2019making,
author = {Ginart, Antonio A. and Guan, Melody Y. and Valiant, Gregory and Zou, James},
title = {Making AI forget you: data deletion in machine learning},
year = {2019},
publisher = {Curran Associates Inc.},
address = {Red Hook, NY, USA},
booktitle = {Proceedings of the 33rd International Conference on Neural Information Processing Systems},
articleno = {316},
numpages = {14}
}

@misc{eu_gdpr_2016,
  title        = {{Regulation (EU) 2016/679 of the European Parliament and of the Council of 27 April 2016 on the protection of natural persons with regard to the processing of personal data and on the free movement of such data (General Data Protection Regulation)}},
  author       = {{European Union}},
  year         = {2016},
  howpublished = {\url{https://eur-lex.europa.eu/eli/reg/2016/679/oj}},
  note         = {Official Journal of the European Union, L 119, 1--88},
}

@book{jain2011introduction,
  title     = {Introduction to Biometrics},
  author    = {Jain, Anil K. and Ross, Arun and Nandakumar, Karthik},
  year      = {2011},
  publisher = {Springer Science \& Business Media},
  isbn      = {978-0-387-77326-1}
}

@inproceedings{taigman2014deepface,
  title     = {DeepFace: Closing the Gap to Human-Level Performance in Face Verification},
  author    = {Taigman, Yaniv and Yang, Ming and Ranzato, Marc'Aurelio and Wolf, Lior},
  booktitle = {Proceedings of the IEEE Conference on Computer Vision and Pattern Recognition (CVPR)},
  pages     = {1701--1708},
  year      = {2014},
  doi       = {10.1109/CVPR.2014.220}
}

@InProceedings{MS_celeb_1m,
    author="Guo, Yandong
    and Zhang, Lei
    and Hu, Yuxiao
    and He, Xiaodong
    and Gao, Jianfeng",
    editor="Leibe, Bastian
    and Matas, Jiri
    and Sebe, Nicu
    and Welling, Max",
    title="MS-Celeb-1M: A Dataset and Benchmark for Large-Scale Face Recognition",
    booktitle="Computer Vision -- ECCV 2016",
    year="2016",
    publisher="Springer International Publishing",
    address="Cham",
    pages="87--102",
    isbn="978-3-319-46487-9"
}

@inproceedings{Krizhevsky2009LearningML,
  title={Learning Multiple Layers of Features from Tiny Images},
  author={Alex Krizhevsky},
  year={2009},
  booktitle={Learning Multiple Layers of Features from Tiny Images},
  url={https://api.semanticscholar.org/CorpusID:18268744}
}

@inproceedings{meng2021magface,
  title     = {MagFace: A Universal Representation for Face Recognition and Quality Assessment},
  author    = {Meng, Qingliang and Zhao, Shichao and Huang, Zhida and Zhou, Feng},
  booktitle = {Proceedings of the IEEE/CVF Conference on Computer Vision and Pattern Recognition (CVPR)},
  pages     = {14225--14234},
  year      = {2021}
}

@inproceedings{deng2019arcface,
  title={Arcface: Additive angular margin loss for deep face recognition},
  author={Deng, Jiankang and Guo, Jia and Xue, Niannan and Zafeiriou, Stefanos},
  booktitle={Proceedings of the IEEE/CVF conference on computer vision and pattern recognition},
  pages={4690--4699},
  year={2019}
}

@INPROCEEDINGS{9880230,
  author={Kim, Minchul and Jain, Anil K. and Liu, Xiaoming},
  booktitle={2022 IEEE/CVF Conference on Computer Vision and Pattern Recognition (CVPR)}, 
  title={AdaFace: Quality Adaptive Margin for Face Recognition}, 
  year={2022},
  volume={},
  number={},
  pages={18729-18738},
  doi={10.1109/CVPR52688.2022.01819}}

@inproceedings{huang_inria-00321923,
  TITLE = {{Labeled Faces in the Wild: A Database for Studying Face Recognition in Unconstrained Environments}},
  AUTHOR = {Huang, Gary B. and Mattar, Marwan and Berg, Tamara and Learned-Miller, Eric},
  URL = {https://inria.hal.science/inria-00321923},
  BOOKTITLE = {{Workshop on Faces in 'Real-Life' Images: Detection, Alignment, and Recognition}},
  ADDRESS = {Marseille, France},
  ORGANIZATION = {{Erik Learned-Miller and Andras Ferencz and Fr{\'e}d{\'e}ric Jurie}},
  YEAR = {2008},
  MONTH = Oct,
  HAL_ID = {inria-00321923},
  HAL_VERSION = {v1},
}

@INPROCEEDINGS{cfpfp2016Sengupta,
  author={Sengupta, Soumyadip and Chen, Jun-Cheng and Castillo, Carlos and Patel, Vishal M. and Chellappa, Rama and Jacobs, David W.},
  booktitle={2016 IEEE Winter Conference on Applications of Computer Vision (WACV)}, 
  title={Frontal to profile face verification in the wild}, 
  year={2016},
  volume={},
  number={},
  pages={1-9},
  doi={10.1109/WACV.2016.7477558}
}

@inproceedings{moschoglou2017agedb,
  title={Agedb: the first manually collected, in-the-wild age database},
  author={Moschoglou, Stylianos and Papaioannou, Athanasios and Sagonas, Christos and Deng, Jiankang and Kotsia, Irene and Zafeiriou, Stefanos},
  booktitle={proceedings of the IEEE conference on computer vision and pattern recognition workshops},
  pages={51--59},
  year={2017}
}

@TechReport{CPLFWTech,
  author =       {T. Zheng and W. Deng},
  title =        {Cross-pose LFW: A database for studying cross-pose face recognition in unconstrained environments},
  institution =  {Beijing University of Posts and Telecommunications},
  year =         {2018},
  number =       {18-01},
  month =        {February}
}

@article{abs-1708-08197,
  author    = {Tianyue Zheng and
               Weihong Deng and
               Jiani Hu},
  title     = {Cross-Age {LFW:} {A} Database for Studying Cross-Age Face Recognition
               in Unconstrained Environments},
  journal   = {CoRR},
  volume    = {abs/1708.08197},
  year      = {2017},
  eprint    = {1708.08197},
}

@InProceedings{Schroff_2015_CVPR,
author = {Schroff, Florian and Kalenichenko, Dmitry and Philbin, James},
title = {FaceNet: A Unified Embedding for Face Recognition and Clustering},
booktitle = {Proceedings of the IEEE Conference on Computer Vision and Pattern Recognition (CVPR)},
month = {June},
year = {2015}
}

@INPROCEEDINGS{cao2015forget,
  author={Cao, Yinzhi and Yang, Junfeng},
  booktitle={2015 IEEE Symposium on Security and Privacy}, 
  title={Towards Making Systems Forget with Machine Unlearning}, 
  year={2015},
  volume={},
  number={},
  pages={463-480},
  doi={10.1109/SP.2015.35}
}

@article{Xu2023survey,
    author = {Xu, Heng and Zhu, Tianqing and Zhang, Lefeng and Zhou, Wanlei and Yu, Philip S.},
    title = {Machine Unlearning: A Survey},
    year = {2023},
    issue_date = {January 2024},
    publisher = {Association for Computing Machinery},
    address = {New York, NY, USA},
    volume = {56},
    number = {1},
    issn = {0360-0300},
    url = {https://doi.org/10.1145/3603620},
    doi = {10.1145/3603620},
    journal = {ACM Comput. Surv.},
    month = aug,
    articleno = {9},
    numpages = {36}
}

@INPROCEEDINGS{MU2021Bourtoule,
  author={Bourtoule, Lucas and Chandrasekaran, Varun and Choquette-Choo, Christopher A. and Jia, Hengrui and Travers, Adelin and Zhang, Baiwu and Lie, David and Papernot, Nicolas},
  booktitle={2021 IEEE Symposium on Security and Privacy (SP)}, 
  title={Machine Unlearning}, 
  year={2021},
  volume={},
  number={},
  pages={141-159},
  doi={10.1109/SP40001.2021.00019}
}

@inproceedings{graves2021amnesiac,
  title={Amnesiac Machine Learning},
  author={Graves, Cody and Nagisetty, Shreyas and Ganesh, Vijay},
  booktitle={Advances in Neural Information Processing Systems (NeurIPS)},
  year={2021}
}

@article{chundawat2023zero, 
title={Zero-shot machine unlearning}, 
author={Chundawat, Vikram S and Tarun, Ayush K and Mandal, Murari and Kankanhalli, Mohan}, 
journal={IEEE Transactions on Information Forensics and Security}, 
year={2023}, 
publisher={IEEE} 
}

@article{Foster_Schoepf_Brintrup_2024,
  title={Fast Machine Unlearning without Retraining through Selective Synaptic Dampening},
  volume={38},
  url={https://ojs.aaai.org/index.php/AAAI/article/view/29092},
  DOI={10.1609/aaai.v38i11.29092},
  number={11},
  journal={Proceedings of the AAAI Conference on Artificial Intelligence},
  author={Foster, Jack and Schoepf, Stefan and Brintrup, Alexandra},
  year={2024},
  month={Mar.},
  pages={12043-12051} 
}

@inproceedings{LFSSD2024Foster,
    author = {Foster, Jack and Schoepf, Stefan and Brintrup, Alexandra},
    year = {2024},
    month = {02},
    pages = {},
    title = {Loss-Free Machine Unlearning},
    doi = {10.48550/arXiv.2402.19308}
}

@INPROCEEDINGS {MIA2017Shokri,
    author = { Shokri, Reza and Stronati, Marco and Song, Congzheng and Shmatikov, Vitaly },
    booktitle = { 2017 IEEE Symposium on Security and Privacy (SP) },
    title = {{ Membership Inference Attacks Against Machine Learning Models }},
    year = {2017},
    volume = {},
    ISSN = {2375-1207},
    pages = {3-18},
    doi = {10.1109/SP.2017.41},
    url = {https://doi.ieeecomputersociety.org/10.1109/SP.2017.41},
    publisher = {IEEE Computer Society},
    address = {Los Alamitos, CA, USA},
    month ={May}
}

@article{MIA2022Hu,
    author = {Hu, Hongsheng and Salcic, Zoran and Sun, Lichao and Dobbie, Gillian and Yu, Philip S. and Zhang, Xuyun},
    title = {Membership Inference Attacks on Machine Learning: A Survey},
    year = {2022},
    issue_date = {January 2022},
    publisher = {Association for Computing Machinery},
    address = {New York, NY, USA},
    volume = {54},
    number = {11s},
    issn = {0360-0300},
    url = {https://doi.org/10.1145/3523273},
    doi = {10.1145/3523273},
    journal = {ACM Comput. Surv.},
    month = sep,
    articleno = {235},
    numpages = {37}
}

@misc{choi2023machine,
    title={Towards Machine Unlearning Benchmarks: Forgetting the Personal Identities in Facial Recognition Systems}, 
    author={Dasol Choi and Dongbin Na},
    journal={AAAI 2024 Workshop on Privacy-Preserving Artificial Intelligence},
    year={2023},
    archivePrefix={arXiv},
}

@inproceedings{Nair2023selective,
    author={Nair, Adithyan M and Kumar, Akshit Sudheer and Kumar, Devakrishna Sanil and T, Anjali},
    year = {2023},
    month = {12},
    pages = {211-216},
    title = {Selective Unlearning in Face Recognition: Forgetting Faces without Compromising Accuracy},
    booktitle={2023 3rd International Conference on Innovative Mechanisms for Industry Applications (ICIMIA)}, 
    doi = {10.1109/ICIMIA60377.2023.10426386}
}

@inproceedings{choi2024towards,
  title={Towards Efficient Machine Unlearning with Data Augmentation: Guided Loss-Increasing (GLI) to Prevent the Catastrophic Model Utility Drop},
  author={Choi, Dasol and Choi, Soora and Lee, Eunsun and Seo, Jinwoo and Na, Dongbin},
  booktitle={Proceedings of the IEEE/CVF Conference on Computer Vision and Pattern Recognition},
  pages={93--102},
  year={2024}
}

@inproceedings{DBLP:conf/eccv/BelabbaciWT26,
  author       = {El Ouanas Belabbaci and
                  Bhavesh Wani and
                  Philipp Terh{\"{o}}rst},
  editor       = {Paolo Favaro and
                  Zuzana Kukelova and
                  Atsuto Maki and
                  Anna Rohrbach and
                  Konrad Schindler and
                  Federico Tombari},
  title        = {Learning to Attract and Repel: Dual Quality Margin Learning for Face
                  Recognition (DQM-Face)},
  booktitle    = {Computer Vision - {ECCV} 2026 - 19th European Conference, Malm{\"{o}},
                  Sweden, September 8-12, 2026, Proceedings, Part {XXVI}},
  series       = {Lecture Notes in Computer Science},
  volume       = {17026},
  pages        = {40--58},
  publisher    = {Springer},
  year         = {2026},
  url          = {https://doi.org/10.1007/978-3-032-37595-7\_3},
  doi          = {10.1007/978-3-032-37595-7\_3},
  bibsource    = {dblp computer science bibliography, https://dblp.org}
}

@InProceedings{Zakharov_2026_WACV,
    author    = {Zakharov, Mikhail},
    title     = {Face Identity Unlearning for Retrieval via Embedding Dispersion},
    booktitle = {Proceedings of the IEEE/CVF Winter Conference on Applications of Computer Vision (WACV) Workshops},
    month     = {March},
    year      = {2026},
    pages     = {1277-1286}
}
}

\end{document}